\documentclass[conference]{IEEEtran}
\IEEEoverridecommandlockouts

\usepackage{cite}
\usepackage{graphicx}
\usepackage{epstopdf}
\usepackage{diagbox}
\usepackage{multirow}
\usepackage[cmex10]{amsmath}
\usepackage[ruled]{algorithm2e} 
\usepackage{url}
\usepackage{lineno,hyperref}
\usepackage{amsthm}
\usepackage{booktabs}
\usepackage{subfigure}
\usepackage{url}
\usepackage{mathrsfs}
\usepackage{amsmath}
\usepackage{amsthm}
\usepackage{bbding}

\usepackage{amsfonts,amssymb}
\usepackage{mathrsfs}
\usepackage{amsmath}
\usepackage{amsthm}
\usepackage{bbding}
\usepackage{cite}
\usepackage{xcolor}
\usepackage{bm}

\usepackage{amsmath,amssymb,amsfonts}
\usepackage{algorithmic}
\usepackage{graphicx}
\usepackage{textcomp}
\usepackage{url}
\usepackage{multirow}
\usepackage{booktabs}
\usepackage{lineno,hyperref}
\usepackage{xcolor}
\def\BibTeX{{\rm B\kern-.05em{\sc i\kern-.025em b}\kern-.08em
    T\kern-.1667em\lower.7ex\hbox{E}\kern-.125emX}}
\begin{document}

\title{Diffusion–Low-Rank Hybrid Reconstruction for Sparse-View Medical Imaging}

\author{\IEEEauthorblockN{Zongyin Deng$^{1}$, Qing Zhou$^{1}$, Yuhao Fang$^{1}$, Zijian Wang$^{2}$, Yao Lu$^{1}$, Ye Zhang$^{1,3}$, and Chun Li$^{1,*}$}
\IEEEauthorblockA{\textit{$^{1}$ Shenzhen MSU-BIT University, Shenzhen, 518172, China.} \\
\textit{$^{2}$China Media Group, Beijing, 100859, China.}\\
\textit{$^{3}$School of Mathematics and Statistics, Beijing Institute of Technology, 100081, Beijing, China.}\\
Email: 575879580@qq.com; zhouqing@smbu.edu.cn; fangyuhao209@gmail.com; wangzijian@cctv.com; vis\_yl@smbu.edu.cn; \\ye.zhang@smbu.edu.cn. 
*Corresponding author: Chun Li (Email: lichun2020@smbu.edu.cn).}
}

\maketitle

\begin{abstract}
        This work presents TV-LoRA, a novel method for low-dose sparse-view CT reconstruction that combines a diffusion generative prior (NCSN++ with SDE modeling) and multi-regularization constraints, including anisotropic TV and nuclear norm (LoRA), within an ADMM framework. To address ill-posedness and texture loss under extremely sparse views, TV-LoRA integrates generative and physical constraints, and utilizes a 2D slice-based strategy with FFT acceleration and tensor-parallel optimization for efficient inference. Experiments on AAPM-2016, CTHD, and LIDC datasets with $N_{\mathrm{view}}=8,4,2$ show that TV-LoRA consistently surpasses benchmarks in SSIM, texture recovery, edge clarity, and artifact suppression, demonstrating strong robustness and generalizability. Ablation studies confirm the complementary effects of LoRA regularization and diffusion priors, while the FFT-PCG module provides a 4.5× speedup. Overall, Diffusion + TV-LoRA achieves high-fidelity, efficient 3D CT reconstruction and broad clinical applicability in low-dose, sparse-sampling scenarios.
\end{abstract}


\section{Introduction}

In recent years, low-dose computed tomography (LDCT) \cite{a1} and sparse-view CT \cite{pp2} have emerged as significant trends in medical imaging, driven by the urgent need to reduce radiation exposure, enhance patient safety, and lower clinical costs. However, when projection views are extremely limited and radiation doses decrease to merely $10\sim20\%$ of standard levels, both model-based iterative reconstruction (MBIR) methods \cite{a3} and purely data-driven deep learning approaches \cite{pp5} face two critical bottlenecks: (1) severely ill-posed imaging processes lead to ill-conditioned inverse problems, resulting in significant loss of details and increased artifacts; and (2) deep learning models are sensitive to training data distribution, potentially producing anatomically implausible structures that undermine clinical utility.

High-quality medical image reconstruction is fundamental for accurate clinical diagnosis and effective treatment planning. However, hardware constraints, limited acquisition time, and reduced radiation exposure often result in medical images with low resolution, high noise, or incomplete data. Consequently, reconstructing high-fidelity images from undersampled or degraded observations remains a significant research challenge.

Traditional methods such as filtered back-projection (FBP), compressed sensing (CS), and deep learning-based models have improved image quality but still face substantial challenges. Iterative methods rely on handcrafted priors, are computationally intensive, and struggle with complex anatomical structures. Deep learning approaches offer computational efficiency but require large, high-quality datasets and risk generating hallucinated details that compromise diagnostic reliability.

Recently, diffusion models have shown significant potential by progressively denoising and synthesizing high-quality medical images. Concurrently, low-rank regularization effectively exploits intrinsic structural similarities in medical images, suppressing noise and recovering missing information. Nevertheless, integrating the generative power of diffusion models with low-rank regularization remains an open research area.

To address these challenges, this study proposes a novel medical image reconstruction method combining diffusion models and low-rank regularization. The main contributions of this work are:

\begin{itemize}
\item \textbf{Integration of diffusion priors with low-rank constraints, leveraging generative capabilities and structural preservation to enhance reconstruction.}
\item \textbf{Development of an adaptive optimization strategy to balance generated detail richness with reconstruction fidelity, mitigating oversmoothing and hallucinated textures.}
\end{itemize}

The proposed method is validated across multiple medical imaging modalities (e.g., MRI, CT) and benchmarked against state-of-the-art techniques, demonstrating superior performance, particularly under low-sampling conditions. This study presents a novel technical paradigm for medical image reconstruction, significantly advancing intelligent medical image analysis.

\renewcommand\arraystretch{1.0}
\begin{table}[t]
\setlength{\belowdisplayskip}{0pt}
\setlength{\abovedisplayskip}{0pt}
\setlength{\abovecaptionskip}{0pt}
\caption{Key notations used in this work. The table summarizes the primary symbols and their definitions, ensuring clarity and consistency throughout the manuscript.}
\centering
\scriptsize
\setlength{\tabcolsep}{2pt}
\begin{tabular}{p{3cm}p{5cm}}
\toprule [1.0pt]
Notation & Definition \\
\midrule[0.5pt]
$x(0)$ & Clean image at initial time \\
$x(T)$ & Maximally corrupted image at terminal time $T$ \\
$\sigma(t)$ & Monotonically increasing noise scale function \\
$\mathrm{d}\mathbf{w}$ & Standard Brownian motion increment \\
$s_\theta(x,t)$ & Time-dependent score function \\
$\mathbf{A}$ & System matrix/forward projection operator \\
$b$ & Measurement vector/projection data \\
$\lambda_{\mathrm{diff}}$ & Weight for diffusion prior term \\
$\alpha$ & Weight for TV regularization \\
$\beta$ & Weight for low-rank regularization \\
$\mathrm{TV}(x)$ & Anisotropic total variation of image $x$ \\
$\|P(x)\|_*$ & Nuclear norm of patch-wise unfolded image \\
$v_x, v_y$ & Auxiliary variables for image gradients \\
$Z$ & Auxiliary variable for low-rank patches \\
$\rho_1, \rho_2, \rho_3$ & ADMM penalty parameters \\
$u_1, u_2, u_3$ & Scaled Lagrange multipliers \\
$N_{\text{view}}$ & Number of projection views \\
$\theta_k$ & $k$-th projection angle ($\theta_k = k\pi/N_{\text{view}}$) \\
$D$ & Number of detector pixels \\
$\sigma_{\min}, \sigma_{\max}$ & Minimum and maximum noise levels for VESDE \\
$N$ & Number of diffusion steps \\
\bottomrule[1.0pt]
\end{tabular}
\label{tab00}
\end{table}

\section{Methodology}

This work proposes a two-stage reconstruction framework that integrates a diffusion-based generative prior with sparsity-promoting physical constraints, enabling robust and high-fidelity image recovery from limited or noisy measurements. The key notations used throughout this paper are summarized in Table \ref{tab00}.

\subsubsection{Overall Framework}
The proposed approach consists of two tightly coupled stages: \textbf{Stage 1:} An initial estimate is generated from pure noise using a diffusion model, specifically by solving an inverse stochastic differential equation (SDE). \textbf{Stage 2:} This estimate is further refined by enforcing both anisotropic total variation (TV) and low-rank (nuclear norm) regularization. The overall joint optimization is efficiently solved via the Alternating Direction Method of Multipliers (ADMM).
Throughout, ADMM iterations are embedded within the diffusion process, balancing data fidelity ($\|\mathbf{A}x - b\|_2^2$) and learned priors, while leveraging GPU parallelization for computational efficiency.

\subsubsection{Diffusion Model: Forward and Reverse Processes}

\textbf{Forward Process (Variance-Exploding SDE):} The forward diffusion process gradually corrupts a clean image $x(0)$ by adding Gaussian noise over time, resulting in a highly noisy sample $x(T)$. This process is formally described by the following SDE: $\mathrm{d}x = \sqrt{\frac{\mathrm{d}\sigma^2(t)}{\mathrm{d}t}}\,\mathrm{d}\mathbf{w}, \quad t \in [0, T],$
where $\sigma(t)$ is a monotonically increasing noise scale function, and $\mathrm{d}\mathbf{w}$ denotes standard Brownian motion. Here, $x(0)$ is the clean image, and $x(T) \sim \mathcal{N}(0, \sigma^2(T)\mathbf{I})$ is the maximally corrupted image at terminal time $T$.

\textbf{Inverse (Reverse) Process:} To generate a clean image from noise, a time-dependent score function $s_\theta(x, t) \approx \nabla_x \log p_t(x)$ is learned, where $p_t(x)$ is the data distribution at time $t$. The reverse-time SDE for denoising is: $\mathrm{d}x = \left[ -\frac{1}{2} \frac{\mathrm{d}\sigma^2(t)}{\mathrm{d}t} \, s_\theta(x, t) \right]\mathrm{d}t
+ \sqrt{\frac{\mathrm{d}\sigma^2(t)}{\mathrm{d}t}}\,\mathrm{d}\bar{\mathbf{w}},$
where $\mathrm{d}\bar{\mathbf{w}}$ is a standard Brownian increment in reverse time.

In practice, this SDE is discretized using the Euler--Maruyama scheme:
\begin{equation}
x_{i-1} = x_i + (\sigma_i^2 - \sigma_{i-1}^2)\, s_\theta(x_i, t_i)
+ \sqrt{\sigma_i^2 - \sigma_{i-1}^2} \cdot z_i,
\end{equation}
where $z_i \sim \mathcal{N}(0, \mathbf{I})$ and the time interval is divided into $N$ steps, $\Delta t = T/N$.

\subsubsection{Joint Optimization Formulation}

The reconstruction is posed as the minimization of the following composite objective:
\begin{equation}
\min_{x}\; \frac{1}{2}\|\mathbf{A}x-b\|_2^2
+ \lambda_{\mathrm{diff}}\bigl[-\log p_\theta(x)\bigr]
+ \alpha\,\mathrm{TV}(x) + \beta\|P(x)\|_*,
\end{equation}
where $\mathbf{A}x \approx b$ enforces data fidelity to measurements $b$, $p_\theta(x)$ is the diffusion generative prior, $\mathrm{TV}(x)$ is anisotropic total variation, and $\|P(x)\|_*$ denotes the nuclear norm (low-rank) of the patch-wise unfolded image.

\subsubsection{ADMM-Based Solution Strategy}

To efficiently handle the non-smooth TV and low-rank regularizations, we introduce auxiliary variables:1. $v_x = \nabla_x x$ and $v_y = \nabla_y x$ (for image gradients), 2. $Z = P(x)$ (for low-rank patches). The constrained optimization problem is reformulated as:
\begin{equation}
\min_{x, v_x, v_y, Z}\;
\frac{1}{2}\|\mathbf{A}x-b\|_2^2 + \alpha(\|v_x\|_1 + \|v_y\|_1) + \beta\|Z\|_*
\end{equation}
\begin{equation*}
\text{subject to:} \;\; v_x = \nabla_x x,\;\; v_y = \nabla_y x,\;\; Z = P(x).
\end{equation*}
The augmented Lagrangian is then constructed, and the solution proceeds via the following ADMM updates:

\subsubsection{ADMM Iterative Updates}
\begin{itemize}
    \item \textbf{Diffusion Denoising Step:} 
    Before each ADMM iteration, the reverse diffusion process is performed to refine $x$ using the score network $s_\theta^*$.

    \item \textbf{Image Variable Update ($x$-step):}
\begin{equation}
\begin{array}{l}
\left( {{{\bf{A}}^T}{\bf{A}} + {\rho _1}\nabla _x^T{\nabla _x} + {\rho _2}\nabla _y^T{\nabla _y} + {\rho _3}I} \right){x^{k + 1}}\\
\begin{array}{*{20}{c}}
{}&{}&{}
\end{array} = {{\bf{A}}^T}b + {\rho _1}\nabla _x^T(v_x^k + u_1^k) + {\rho _2}\nabla _y^T(v_y^k + u_2^k)\\
\begin{array}{*{20}{c}}
{}&{}&{}
\end{array} + {\rho _3}({Z^k} + u_3^k),
\end{array}
\end{equation}
    where $\rho_i$ are penalty parameters and $u_i$ are scaled Lagrange multipliers.

    \item \textbf{Gradient Variable Update ($g$-step):}
    \begin{equation}
    v \leftarrow \operatorname{sign}(v) \odot \max(|v| - \tau, 0),
    \end{equation}
    where $\tau = \alpha/\rho$.

    \item \textbf{Low-Rank Variable Update ($z$-step):}
    For each patch matrix, update $Z$ via singular value thresholding (SVT):
    \begin{equation}
    \sigma_i \leftarrow \max(\sigma_i - \beta/\rho, 0),
    \end{equation}
    and reconstruct $Z = U\tilde{\Sigma}V^T$.

    \item \textbf{Lagrange Multiplier Update ($U$-step):}
\begin{equation}
\begin{aligned}
u_1^{k+1} &= u_1^k + \bigl(v_x^{k+1} - \nabla_x x^{k+1}\bigr),\\
u_2^{k+1} &= u_2^k + \bigl(v_y^{k+1} - \nabla_y x^{k+1}\bigr),\\
u_3^{k+1} &= u_3^k + \bigl(Z^{k+1} - P(x^{k+1})\bigr).
\end{aligned}
\end{equation}
\end{itemize}
This procedure is repeated until convergence. The overall energy decreases monotonically and, under mild convexity assumptions, the method converges with a rate of $\mathcal{O}(1/k)$.

By embedding the diffusion generative prior into the ADMM framework and explicitly balancing measurement consistency, noise suppression, edge preservation, and global low-rank structure, the proposed methodology achieves superior, diagnostically reliable image reconstructions, particularly under challenging, incomplete, or noisy data conditions.
\begin{algorithm}[t]
    \caption{\small Diffusion–Low-Rank Hybrid Reconstruction for Sparse-View Medical Imaging}
    \label{alg:diff_sparse_recon}
    \DontPrintSemicolon
    \small
    \textbf{Requires:} $s_\theta^{*},\; b,\; A,\; \alpha,\; \beta,\; \rho_1,\; \rho_2,\; \rho_3,\; \mathbf{N}$\\
    \textbf{Define:} $D_x,\; D_y \leftarrow$ Gradient operators in $x$/$y$ directions\\
    \textbf{Define:} $P(\cdot) \leftarrow$ Patch-unfolding operator that maps an image to a block matrix\\
    \textbf{Define:} $\operatorname{Shrink}(v,\tau) \leftarrow \operatorname{sign}(v)\!\odot\!\max(|v|-\tau,0)$ (soft threshold)\\
    \textbf{Define:} $\operatorname{SVT}(M,\tau)\!:\! M=U\Sigma V^{\!\top},\; U\!\operatorname{diag}(\max(\sigma_i-\tau,0))V^{\!\top}$\\
    $x^{0}\!\sim\!\mathcal{N}(\mathbf{0},\sigma_{T}^{2}\boldsymbol{I});\;
     v_x^{0}\!\gets\!0;\;
     v_y^{0}\!\gets\!0;\;
     Z^{0}\!\gets\!0;\;
     u_1^{0}\!\gets\!0;\;
     u_2^{0}\!\gets\!0;\;
     u_3^{0}\!\gets\!0.$\\

    \For{$k = 0$ to $\mathbf{N}-1$}{
        \tcp*[f]{\textbf{*Diffusion denoise*}}\\
        ${x^{k}}^{\prime} \gets \mathrm{Solve}(x^{k},\, s_\theta^{*})$\\[2pt]

        \tcp*[f]{\textbf{*$x$ step*}}\\
        $A_{\!x} \gets A^{\!\top}A + \rho_1 D_x^{\!\top}D_x + \rho_2 D_y^{\!\top}D_y + \rho_3 I$\\
        $b_{\!x} \gets A^{\!\top}b + \rho_1 D_x^{\!\top}(v_x^{k} + u_1^{k})
                          + \rho_2 D_y^{\!\top}(v_y^{k} + u_2^{k})
                          + \rho_3 (Z^{k} + u_3^{k})$\\
        $x^{k+1} \gets \mathrm{PCG}(A_{\!x},\, b_{\!x},\, {x^{k}}^{\prime},\, 1)$\\[2pt]

        \tcp*[f]{\textbf{*$v$ step* (TV shrinkage)}}\\
        $v_x^{k+1} \gets \operatorname{Shrink}\!\bigl(D_x x^{k+1} - u_1^{k},\, \alpha/\rho_1\bigr)$\\
        $v_y^{k+1} \gets \operatorname{Shrink}\!\bigl(D_y x^{k+1} - u_2^{k},\, \alpha/\rho_2\bigr)$\\[2pt]

        \tcp*[f]{\textbf{*$Z$ step* (low-rank SVT)}}\\
        $X^{k+1} \gets P(x^{k+1})$\\
        $Z^{k+1} \gets \operatorname{SVT}\!\bigl(X^{k+1} - u_3^{k},\, \beta/\rho_3\bigr)$\\[2pt]

        \tcp*[f]{\textbf{*$u$ step* (multipliers)}}\\
        $u_1^{k+1} \gets u_1^{k} + v_x^{k+1} - D_x x^{k+1}$\\
        $u_2^{k+1} \gets u_2^{k} + v_y^{k+1} - D_y x^{k+1}$\\
        $u_3^{k+1} \gets u_3^{k} + Z^{k+1} - X^{k+1}$\\
    }
    \textbf{Output:} reconstructed image $x^{N}$
\end{algorithm}

\section{Experiments}
\subsubsection{Datasets} To evaluate the generalizability and robustness of our method across diverse scanning protocols and lesion types, we use three public CT datasets: LDCT (AAPM 2016) \cite{bu3}, CTHD \cite{bu4}, and LIDC-IDRI \cite{bu5}. LDCT provides abdominal scans with both low- and standard-dose images from 198 patients, formatted as $256 \times 256$ slices and normalized to $[0, 1]$. CTHD focuses on liver lesion detection with expert-annotated 3D volumes, also preprocessed to $256 \times 256$ pixels with preserved gray-levels. LIDC-IDRI contains thousands of annotated thoracic CT scans for lung nodule analysis, center-cropped and rescaled for consistency. These datasets collectively cover abdominal, hepatic, and pulmonary CT tasks, ensuring broad applicability of our approach.

\subsubsection{Metrics} For objective assessment of image reconstruction performance, we employ two standard quantitative metrics: Peak Signal-to-Noise Ratio (PSNR) \cite{pp31} and Structural Similarity Index Measure (SSIM) \cite{pp31}. PSNR quantifies pixel-level reconstruction fidelity by measuring the ratio between the maximum possible signal and the reconstruction error. SSIM evaluates perceptual image quality by considering luminance, contrast, and structural similarity, using parameters suitable for 16-bit medical images. Together, these metrics provide a comprehensive and robust evaluation of reconstruction quality.

\subsubsection{Software and Hardware Environment} All experiments were conducted on a high-performance computing platform. The hardware configuration included an Intel Xeon 8358P processor with 32 cores and 64 threads at 2.6 GHz, and two NVIDIA RTX 4090 GPUs, each with 24 GB GDDR6X memory and a single-precision computing capability of 8.9 TFLOPS. The software environment was based on Ubuntu 22.04 LTS with Linux Kernel 5.15. The computational framework utilized CUDA 11.8, cuDNN 8.7, and PyTorch 2.0.1~\cite{bu6}, enabling efficient execution of large-scale 3D image parallel reconstruction tasks. The Adam \cite{pp32} optimizer is employed for all model training procedures.

\subsubsection{Data Processing} The data preprocessing pipeline comprises the following steps. First, a Radon transform is applied to the 3D volumetric data to generate $180^{\circ}$ full-angle projection data. The number of detector pixels is set to $D = \lceil 256\sqrt{2} \rceil$ to ensure complete sampling under rotational symmetry. For a specified number of projection views $N_{\text{view}} \in {8, 4, 2}$, projection angles are uniformly sampled within the range $[0, \pi]$ as $\theta_k = k\pi/N_{\text{view}}$. Based on these selected angles, sparse-view projection measurement vectors $\mathbf{b}$ and their corresponding system matrices $\mathbf{A}$ are generated and stored. To enhance model robustness, Gaussian noise $\mathcal{N}(0, \sigma^2)$ may be added in the projection domain to simulate uncertainties inherent in practical low-dose acquisitions.

\subsubsection{Implementation Details} The diffusion prior module (NCSN++) adopts a four-level U-Net~\cite{qq32} as its backbone, with channel dimensions of 128, 128, 256, and 512, respectively. Each level comprises a \textit{Conv–GroupNorm–SiLU} double convolution block and inter-level skip connections, which collectively enhance feature representation and multi-scale information fusion. At the lowest level, a global attention mechanism (GEGLU) is incorporated to further improve the modeling of complex textures and long-range dependencies.

For noise scheduling, we employ the Variance Exploding SDE (VESDE) with parameters $\sigma_{\min}=0.01$ and $\sigma_{\max}=50$, using a total of $N=2000$ steps. A log-uniform sampling strategy is utilized along the time dimension to ensure comprehensive coverage and high resolution within the noise space. Model training is conducted via denoising score matching combined with a multi-scale $\ell_2$ residual loss, which enhances the model’s ability to capture image details at various scales compared to the original NCSN++.

Optimization is performed using the Adam optimizer~\cite{pp32} with an initial learning rate of $1\times 10^{-5}$, $\beta_1=0.9$, and $\beta_2=0.999$. Cosine annealing is employed to mitigate overfitting and accelerate convergence. During training, model parameters are updated using an exponential moving average (EMA) with a decay rate of 0.999. Only EMA-averaged weights are used during inference to ensure model stability and generalization.

The model comprises approximately 43.7 million parameters and supports efficient training with a batch size of 12 on a single RTX 4090 (24GB) GPU. This configuration achieves a strong balance between representational capacity and computational efficiency, making it well-suited for large-scale 3D medical image reconstruction tasks.

\begin{table}
\centering
\caption{Quantitative evaluation of SV-CT reconstruction quality under various hyperparameter settings $(N_{\mathrm{view}},\,\rho_0,\,\rho_1)$ on the LDCT $256 \times 256$ test set.}
\label{tab:1}
\setlength{\tabcolsep}{6pt}
\begin{tabular}{lcccccc}
\toprule
\multirow{2}{*}{Setting}
 & \multicolumn{2}{c}{Axial} & \multicolumn{2}{c}{Coronal} & \multicolumn{2}{c}{Sagittal}\\
\cmidrule(lr){2-3}\cmidrule(lr){4-5}\cmidrule(lr){6-7}
 & PSNR$\uparrow$ & SSIM$\uparrow$
 & PSNR$\uparrow$ & SSIM$\uparrow$
 & PSNR$\uparrow$ & SSIM$\uparrow$\\
\midrule
$(8,5,5)$   & 31.04 & 0.894 & 32.68 & 0.890 & 31.62 & 0.894\\
$(8,10,1)$  & \textbf{31.38} & \textbf{0.906} & \textbf{33.27} & \textbf{0.905} & \textbf{32.03} & \textbf{0.907}\\
$(8,10,3)$  & 27.84 & 0.903 & 30.43 & 0.903 & 28.62 & 0.904\\
\midrule
$(4,5,5)$   & 27.11 & 0.768 & 27.15 & 0.754 & 27.08 & 0.763\\
$(4,10,1)$  & 29.07 & 0.840 & 29.41 & 0.832 & 29.15 & 0.837\\
$(4,10,3)$  & \textbf{29.56} & \textbf{0.842} & \textbf{29.81} & \textbf{0.834} & \textbf{29.59} & \textbf{0.839}\\
\midrule
$(2,5,5)$   & 26.32 & 0.800 & 28.08 & 0.790 & 26.64 & 0.796\\
$(2,10,1)$  & \textbf{26.83} & 0.777 & \textbf{27.28} & 0.764 & \textbf{26.78} & 0.770\\
$(2,10,3)$  & 26.34 & \textbf{0.789} & 27.20 & \textbf{0.779} & 26.36 & \textbf{0.784}\\
\bottomrule
\end{tabular}
\end{table}

\begin{table}
\centering
\caption{Optimal ADMM–TV-LoRA hyperparameter settings for different numbers of views (2, 4, and 8). This table summarizes the specific hyperparameter configurations that yield the best performance for each view setting, highlighting how the optimal parameters vary with the number of perspectives in the reconstruction task.}
\label{tab:2}
\renewcommand{\arraystretch}{1.5}
\setlength{\tabcolsep}{10pt}
\begin{tabular}{cccccc}
\toprule
$N_{\text{view}}$ & $\rho_0$ & $\rho_1$ & $\lambda_{\mathrm{TV}}$ & $\lambda_{\mathrm{LoRA}}$ & remark\\
\midrule
8 & 10 & 1 & 0.04 & 0.02 & \texttt{n\_view8}\\
4 & 10 & 3 & 0.04 & 0.02 & \texttt{n\_view4}\\
2 & 10 & 3 & 0.04 & 0.02 & \texttt{n\_view2}\\
\bottomrule
\end{tabular}
\end{table}

\begin{table}
\centering
\caption{Quantitative evaluation of the TV-LoRA method on LDCT, CTHD, and LIDC datasets under optimal settings $(N_{\mathrm{view}}, \rho_0, \rho_1)$.}
\label{tab:3}
\setlength{\tabcolsep}{3.6pt}
\begin{tabular}{llcccccc}
\toprule
dataset & Setting & \multicolumn{2}{c}{Axial} & \multicolumn{2}{c}{Coronal} & \multicolumn{2}{c}{Sagittal}\\
\cmidrule(lr){3-4} \cmidrule(lr){5-6} \cmidrule(lr){7-8}
& & PSNR$\uparrow$ & SSIM$\uparrow$ & PSNR$\uparrow$ & SSIM$\uparrow$ & PSNR$\uparrow$ & SSIM$\uparrow$\\
\midrule
LDCT & $(8,10,1)$ & 31.38 & \textbf{0.906} & 33.27 & \textbf{0.905} & 32.03 & \textbf{0.907} \\
     & $(4,10,3)$ & 29.56 & \textbf{0.842} & 29.81 & 0.834 & 29.59 & 0.839 \\
     & $(2,10,1)$ & 26.83 & 0.777 & 27.28 & 0.764 & 26.78 & 0.770 \\
\midrule
CTHD & $(8,10,1)$ & \textbf{33.32} & 0.853 & \textbf{34.18} & 0.864 & \textbf{33.51} & 0.867 \\
     & $(4,10,3)$ & \textbf{31.46} & 0.832 & \textbf{32.86} & \textbf{0.841} & \textbf{31.75} & \textbf{0.843} \\
     & $(2,10,3)$ & \textbf{28.46} & \textbf{0.794} & \textbf{30.27} & \textbf{0.797} & \textbf{28.76} & \textbf{0.800} \\
\midrule
LIDC & $(8,10,1)$ & 25.42 & 0.626 & 25.94 & 0.662 & 25.49 & 0.659\\
     & $(4,10,3)$ & 21.67 & 0.545 & 22.20 & 0.581 & 21.71 & 0.571\\
     & $(2,10,3)$ & 18.68 & 0.465 & 19.01 & 0.479 & 18.61 & 0.468\\

\bottomrule
\end{tabular}
\end{table}

\subsubsection{Method Comparison} To comprehensively evaluate the performance of the proposed method, we compare it against seven representative baseline models, encompassing traditional regularization-based techniques, deep learning approaches, and diffusion–optimization hybrid methods. The selected baselines are described as follows: \textbf{(1) ADMM-TGV, SIAM JIS 2010}~\cite{a18}: These methods utilize higher-order Total Generalized Variation (TGV) regularization, imposing smoothness constraints along the axial ($z$) and transverse ($xy$) directions, respectively. Such regularization is particularly well-suited for capturing multi-scale structural features in medical images. \textbf{(2) ADMM-TV, IEEE TMI 2012}~\cite{bu7}: A classical baseline for sparse reconstruction that employs anisotropic Total Variation (TV) as a sparsity prior and solves the associated optimization problem using the Alternating Direction Method of Multipliers (ADMM). This approach effectively suppresses noise while preserving sharp image edges. \textbf{(3) FBPConvNet, IEEE TIP 2017}~\cite{a17}: An end-to-end reconstruction framework that integrates the classical Filtered Back Projection (FBP) algorithm with a U-Net architecture. The FBP algorithm generates an initial reconstruction, which is subsequently refined by a convolutional neural network, providing both high reconstruction speed and strong learning capability. \textbf{(4) Chung \emph{et al., NeurIPS 2022}}~\cite{bu8}: A score-based diffusion reconstruction model for inverse problems, which embeds a noise estimator in the measurement space. This enables direct inversion for image recovery and represents an early application of score-driven generative models in the field. \textbf{(5) Lahiri \emph{et al., CVPR 2023}}~\cite{bu9}: This work introduces ConRad, a structure-aware conditional diffusion model that jointly embeds noise modeling and data consistency. It supports conditional image generation across different modalities and structures, demonstrating robustness in scenarios with missing modalities.

\begin{figure}
    \centering
    \includegraphics[width=\linewidth]{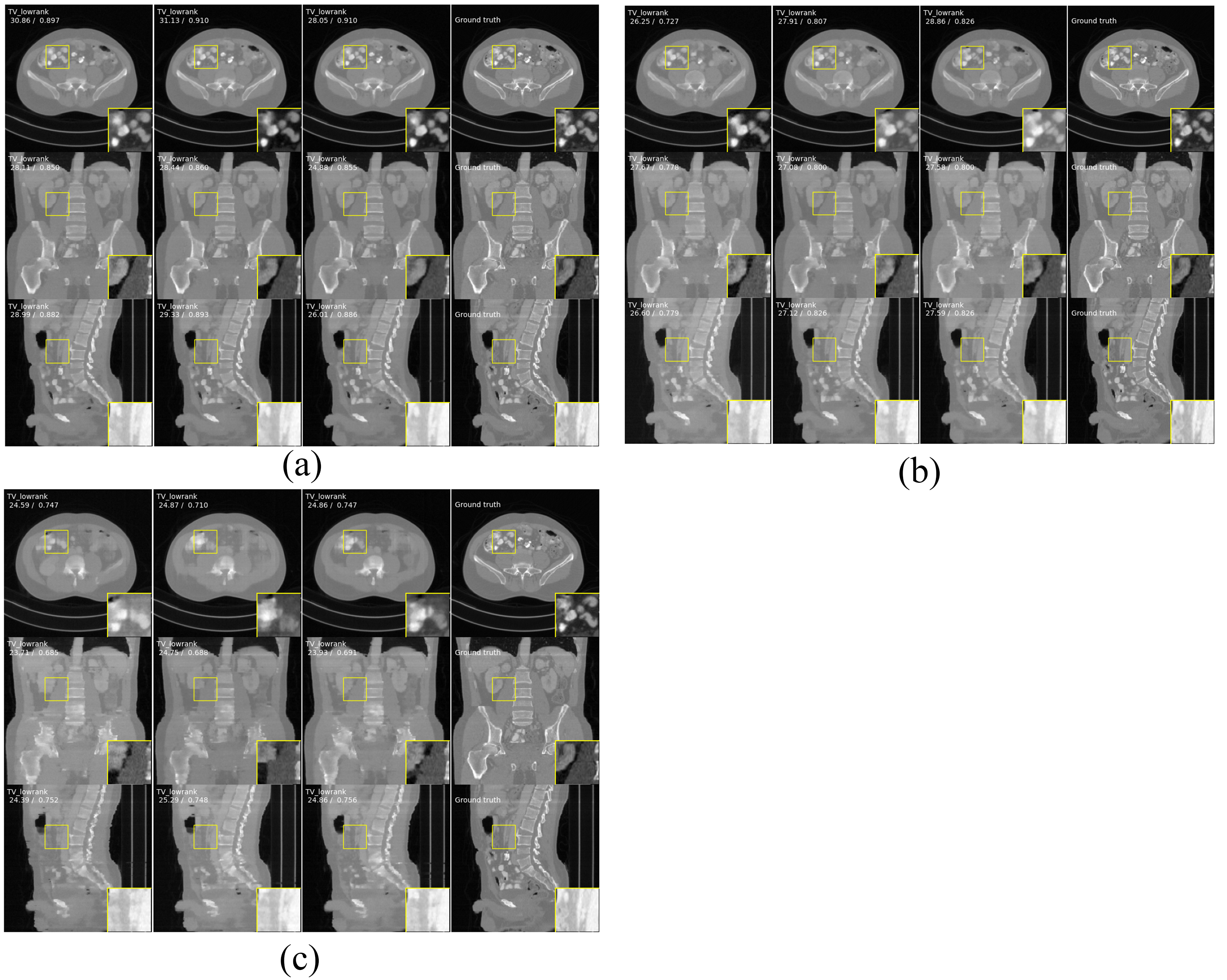}
    \caption{Quantitative comparison of sparse-view CT (SV-CT) reconstruction results on the LDCT dataset with $N_{\mathrm{view}} = 8, 4, 2$. (a) 8 views, (b) 4 views, (c) 2 views. The results demonstrate that the proposed method consistently preserves structural integrity and fine details across different sparsity levels, outperforming baseline approaches.}
    \label{fig:a}
\end{figure}
\begin{figure*}
    \centering
    \includegraphics[width=\linewidth]{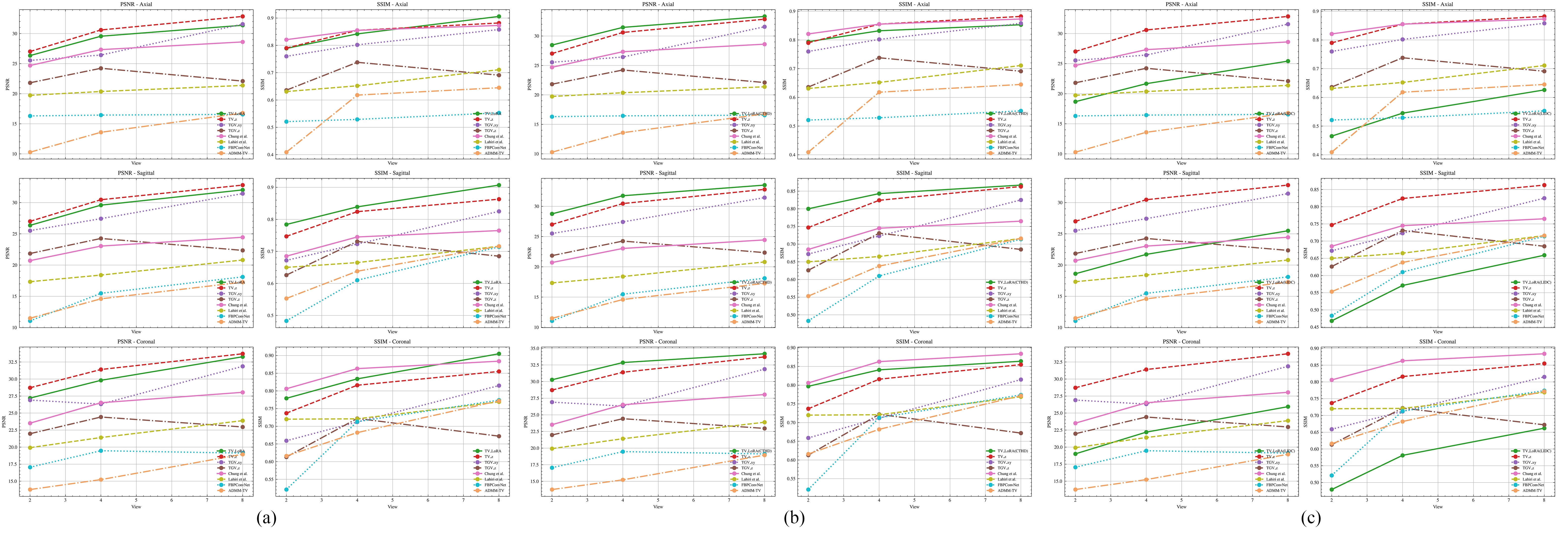}
    \caption{Comparison of TV-LoRA and seven representative baseline models on the (a) LDCT, (b) CTHD, and (c) LIDC datasets in terms of PSNR and SSIM for three orthogonal slices under varying numbers of projection views ($N_{\mathrm{view}}=2, 4, 8$). }
    \label{fig:b}
\end{figure*}

\begin{table}[t]
    \centering
    \small
    \caption{Quantitative evaluation of SV-CT (2, 4, and 8 views) on the CTHD $256 \time 256$ test set. Comparisons are made between the Axial* plane, where the diffusion model prior is applied, and the Coronal and Sagittal views, all measured against the original images.}
    \resizebox{0.5\textwidth}{!}{
        \begin{tabular}{l|ccccccc}
            \toprule[1pt]
            \multirow{2}{*}{View} & \multirow{2}{*}{Method} & \multicolumn{2}{c}{Axial*} & \multicolumn{2}{c}{Coronal} & \multicolumn{2}{c}{Sagittal} \\
            \cmidrule(lr){3-4} \cmidrule(lr){5-6} \cmidrule(lr){7-8}
            & & PSNR $\uparrow$ & SSIM $\uparrow$ & PSNR $\uparrow$ & SSIM $\uparrow$ & PSNR $\uparrow$ & SSIM $\uparrow$ \\
            \midrule[0.5pt]
            \multirow{8}{*}{8 view} & ADMM-TV~\cite{bu7} & 18.05 & 0.650 & 19.40 & 0.775 & 17.70 & 0.717\\
            & TGV\textsubscript{$z$}~\cite{a18}   & 24.35 & 0.725 & 25.18 & 0.707 & 24.60 & 0.718\\
            & TGV\textsubscript{$xy$}~\cite{a18}  & 32.05 & 0.844 & 32.32 & 0.804 & 31.88 & 0.816\\
            & Chung \emph{et al.}\cite{bu8} & 29.45 & \underline{0.874} & 28.92 & \textbf{0.886} & 25.10 & 0.771\\
            & Lahiri \emph{et al.}\cite{bu9} & 22.53 & 0.726 & 24.13 & 0.778 & 21.70 & 0.725\\
            & FBPConvNet~\cite{a17} & 17.85 & 0.560 & 20.25 & 0.780 & 18.75 & 0.719\\
            & TV\textsubscript{$z$}  & \underline{33.05} & \textbf{0.880} & \underline{33.55} & 0.851 & \underline{33.03} & \underline{0.862} \\
            & TV\_LoRA  & \textbf{33.32} & 0.853 & \textbf{34.18} & \underline{0.864} & \textbf{33.51} & \textbf{0.867} \\
            \midrule[0.5pt]
            \multirow{8}{*}{4 view} & ADMM-TV~\cite{bu7} & 14.83 & 0.629 & 16.87 & 0.699 & 15.28 & 0.670\\
            & TGV\textsubscript{$z$}~\cite{a18}      & 24.67 & 0.741 & 25.12 & 0.724 & 24.85 & 0.740\\
            & TGV\textsubscript{$xy$}~\cite{a18}     & 27.12 & 0.815 & 26.99 & 0.783 & 27.56 & 0.800\\
            & Chung \emph{et al.}\cite{bu8} & 28.42 & \textbf{0.858} & 28.16 & \textbf{0.864} & 24.30 & 0.755\\
            & Lahiri \emph{et al.}\cite{bu9} & 20.68 & 0.664 & 21.82 & 0.733 & 19.07 & 0.679\\
            & FBPConvNet~\cite{a17} & 17.28 & 0.577 & 19.56 & 0.749 & 17.02 & 0.662\\
            & TV\textsubscript{$z$} & \underline{30.77} & \underline{0.849} & \underline{31.25} & 0.818 & \underline{30.54} & \underline{0.823} \\
            & TV\_LoRA  & \textbf{31.46} & 0.832 & \textbf{32.86} & \underline{0.841} & \textbf{31.75} & \textbf{0.843} \\
            \midrule[0.5pt]
            \multirow{8}{*}{2 view} & ADMM-TV~\cite{bu7} & 12.65 & 0.479 & 14.43 & 0.602 & 13.04 & 0.532\\
            & TGV\textsubscript{$z$}~\cite{a18}      & 22.11 & 0.657 & 22.91 & 0.634 & 22.60 & 0.645\\
            & TGV\textsubscript{$xy$}~\cite{a18}     & 25.88 & 0.765 & 26.75 & 0.675 & 25.90 & 0.682\\
            & Chung \emph{et al.}\cite{bu8} & 24.88 & \textbf{0.801} & 23.70 & \underline{0.785} & 21.15 & 0.715\\
            & Lahiri \emph{et al.}\cite{bu9}  & 20.47 & 0.662 & 21.60 & 0.740 & 19.10 & 0.670\\
            & FBPConvNet~\cite{a17} & 17.11 & 0.546 & 18.65 & 0.563 & 15.32 & 0.487\\
            & TV\textsubscript{$z$} & \underline{27.35} & \underline{0.798} & \underline{28.40} & 0.725 & \underline{27.30} & \underline{0.736} \\
            & TV\_LoRA  & \textbf{28.46} & 0.794 & \textbf{30.27} & \textbf{0.797} & \textbf{28.76} & \textbf{0.800} \\
            \bottomrule[1.0pt]
        \end{tabular}
    }
    \label{tab:compare_CTHD}
\end{table}

\subsubsection{Results Analysis}

Tables~\ref{tab:1}--\ref{tab:compare_CTHD} sequentially summarize the
quantitative performance of TV\mbox{-}LoRA and the competing
methods at three projection counts ($N_{\text{view}} = 8, 4, 2$),
following the order
\emph{LDCT hyper-parameter search $\rightarrow$ optimal hyper-parameters
$\rightarrow$ cross-dataset comparison $\rightarrow$ detailed CTHD
comparison}: %
Table~\ref{tab:1} first reports axial, coronal, and sagittal
\mbox{PSNR/SSIM} scores for TV\mbox{-}LoRA on LDCT across nine
$(N_{\text{view}},\rho_{0},\rho_{1})$ settings; %
Table~\ref{tab:2} then distills the optimal
\mbox{ADMM--TV-LoRA} hyper-parameters
$(\rho_{0},\rho_{1},\lambda_{\mathrm{TV}},\lambda_{\mathrm{LoRA}})$
for each projection count; %
Table~\ref{tab:3} applies these optimal configurations to compare
three-view reconstruction quality for TV\mbox{-}LoRA on the LDCT, CTHD,
and LIDC datasets; %
finally, Table~\ref{tab:compare_CTHD} uses the latest experimental data
to give a detailed head-to-head comparison between TV\mbox{-}LoRA and
seven baselines on CTHD, reporting the average \mbox{PSNR} and
\mbox{SSIM} over the three anatomical planes.

Figure~\ref{fig:a} visually shows how TV\mbox{-}LoRA preserves fine
structures on LDCT as the number of views decreases from 8 to 2:
even at $N_{\text{view}} = 2$ , soft-tissue
boundaries and subtle textures remain clear, whereas the baselines
display noticeable blur or streak artefacts.  Figure~\ref{fig:b}
plots the PSNR/SSIM curves for seven methods across the three datasets
and view counts, confirming that TV\mbox{-}LoRA consistently occupies
the high-quality region in all slice directions.

\textbf{High sampling rate ($N_{\text{view}} = 8$).}  
On CTHD, TV\mbox{-}LoRA attains an average three-view
\mbox{PSNR/SSIM} of \textbf{33.67 dB/0.861}.  It delivers the highest
PSNR (34.18 dB, 33.51 dB) and SSIM (0.864, 0.867) on the coronal and
sagittal planes, respectively.  The axial PSNR is likewise best
(33.32 dB), although the axial SSIM (0.853) is slightly lower than
that of TV\textsubscript{$z$} (0.880).  Overall, TV\mbox{-}LoRA leads
in PSNR on all three directions, with only a marginal SSIM deficit on
the axial plane, demonstrating sharp detail and minimal numerical
error at high sampling rates.

\textbf{Moderate sampling rate ($N_{\text{view}} = 4$).}  
TV\mbox{-}LoRA achieves \textbf{32.02 dB/0.839} on CTHD,
exceeding TV\textsubscript{$z$} by 1.25 dB and 0.009 in
PSNR and SSIM, respectively, and surpassing every iterative or
deep-learning baseline.  Compared with FBPConvNet, the PSNR gain exceeds
\textbf{15 dB}; Fig.\,\ref{fig:a} further shows a marked reduction in
over-smoothing and “plastic” texture artefacts.

\textbf{Extreme sparse sampling ($N_{\text{view}} = 2$).}  
With only two projections retained, TV\mbox{-}LoRA still records an
average \textbf{29.16 dB/0.797}, beating
TV\textsubscript{$z$} by 1.81 dB and 0.044 in PSNR and SSIM.
Relative to the deep-learning baselines, TV\mbox{-}LoRA preserves
sharper structures and produces fewer streak artefacts
(Fig.\,\ref{fig:b}).  

Reducing the view count from 8 to 2 decreases TV\mbox{-}LoRA’s average
PSNR by only \textbf{4.5 dB} and SSIM by \textbf{0.064}—far less than
the drops observed for TV\textsubscript{$z$}, FBPConvNet, and ADMM-TV,
highlighting its robustness under severe data scarcity.

In summary, TV\mbox{-}LoRA delivers state-of-the-art reconstruction
quality at high sampling rates and maintains superior structural
consistency and artefact suppression at medium and low sampling
rates.  These results validate the
broad applicability and effectiveness of its synergistic
\emph{diffusion prior + low-rank modelling + TV regularization}
framework under diverse sampling conditions and real-world clinical
scenarios.

\subsubsection{Ablation Study} To quantitatively evaluate the contribution of each key component within the TV-LoRA framework, we conducted an ablation study with $N_{\text{view}}=8$. Four model variants were assessed: (1) the full model (Diffusion + ADMM + TV-LoRA + PCG--FFT); (2) without LoRA ($\beta=0$, retaining only anisotropic TV); (3) without the diffusion prior (using only Gaussian initialization); and (4) without PCG--FFT (solving the linear system without frequency-domain acceleration, using standard CG instead). All other training and inference configurations were kept consistent with the main experiments. The models were compared based on the average PSNR and SSIM across three orientations.

The experimental results show that removing LoRA resulted in a 0.034 decrease in SSIM, indicating that global texture continuity strongly depends on the LoRA-based block low-rank constraint. Excluding the diffusion prior led to drops of 2.27 dB in PSNR and 0.045 in SSIM, along with a noticeable increase in streak artifacts, demonstrating the critical role of the generative prior in noise suppression and structural fidelity. Disabling PCG--FFT had minimal effect on reconstruction quality metrics but significantly increased the per-iteration runtime from 23 ms to 103 ms, highlighting its primary contribution to computational efficiency.

In summary, both the diffusion prior and the LoRA-based nuclear norm constraint are essential for enhancing reconstruction quality, while PCG--FFT acceleration ensures practical computational efficiency. Collectively, these three components constitute the core strengths of the TV-LoRA framework.
\section{Conclusion}
This work addresses high-quality medical image reconstruction from extremely sparse and low-dose data. We propose TV-LoRA, a novel framework that integrates diffusion generative models with anisotropic TV and low-rank LoRA regularizations within an iterative ADMM scheme. This approach jointly models local edges and global textures, substantially reducing artifacts and restoring anatomical details, even with minimal projections. We theoretically establish $\mathcal{O}(1/k)$ convergence for the ADMM–TV-LoRA algorithm under suitable parameters. FFT acceleration and GPU parallelization enable efficient large-scale 3D CT reconstruction. Extensive experiments on three real CT datasets (LDCT, CTHD, LIDC) demonstrate that TV-LoRA consistently surpasses seven state-of-the-art baselines, particularly under highly sparse conditions. Ablation studies further confirm the complementary roles of diffusion priors and low-rank regularization. Overall, TV-LoRA provides theoretical, algorithmic, and practical advances for sparse CT reconstruction. Future work will extend this approach to multi-modal, dynamic, and higher-dimensional imaging problems.

\textbf{Acknowledgement:} This research was supported by the National Key Research and Development Program of China (No. 2022YFC3310300), Guangdong Basic and Applied Basic Research Foundation (No. 2024A1515011774), the National Natural Science Foundation of China (No. 12171036), Shenzhen Sci-Tech Fund (Grant No. RCJC20231211090030059), and Beijing Natural Science Foundation (No. Z210001).

\ifCLASSOPTIONcaptionsoff
\newpage
\fi
	
\bibliographystyle{IEEEtran}
\bibliography{IEEEabrv,references.bib}

\end{document}